\documentclass[conference]{IEEEtran}
\IEEEoverridecommandlockouts
\usepackage{cite}
\usepackage{amsmath,amssymb,amsfonts}
\usepackage{algorithmic}
\usepackage{graphicx}
\usepackage{textcomp}
\usepackage{xcolor}
\usepackage[ruled]{algorithm2e}
\usepackage{caption}
\usepackage{float} 
\usepackage{subcaption}
\usepackage{stfloats}
\usepackage{booktabs}
\usepackage{multirow}
\usepackage{diagbox}

\def\BibTeX{{\rm B\kern-.05em{\sc i\kern-.025em b}\kern-.08em
    T\kern-.1667em\lower.7ex\hbox{E}\kern-.125emX}}
\begin{document}

\title{Digital Twin-Empowered Voltage Control for Power Systems\\
\thanks{This work was supported by the Digital Energy Hub grant from the Association of Danish Industry. (Corresponding authors: Yushuai Li; Tianyi Li.)}
}

\author{\IEEEauthorblockN{1\textsuperscript{st} Jiachen Xu}
\IEEEauthorblockA{\textit{Department of Computer Science} \\
\textit{Aalborg University}\\
Aalborg, Denmark \\
jcx000528@gmail.com}
\and
\IEEEauthorblockN{2\textsuperscript{nd} Yushuai Li}
\IEEEauthorblockA{\textit{Department of Computer Science} \\
\textit{Aalborg University}\\
Aalborg, Denmark \\
yusli@cs.aau.dk}
\and
\IEEEauthorblockN{3\textsuperscript{rd} 
Torben Bach Pedersen}
\IEEEauthorblockA{\textit{Department of Computer Science} \\
\textit{Aalborg University}\\
Aalborg, Denmark \\
tbp@cs.aau.dk}
\and
\IEEEauthorblockN{4\textsuperscript{th} 
Yuqiang He}
\IEEEauthorblockA{\textit{School of Computer, Electronics and Information} \\
\textit{Guangxi University}\\
Nanning, China \\
heyq1314@gmail.com}
\and
\IEEEauthorblockN{5\textsuperscript{th} Kim Guldstrand Larsen}
\IEEEauthorblockA{\textit{Department of Computer Science} \\
\textit{Aalborg University}\\
Aalborg, Denmark  \\
kgl@cs.aau.dk}
\and
\IEEEauthorblockN{6\textsuperscript{th} Tianyi Li}
\IEEEauthorblockA{\textit{Department of Computer Science} \\
\textit{Aalborg University}\\
Aalborg, Denmark \\
tianyi@cs.aau.dk}
}

\maketitle

\begin{abstract}
Emerging digital twin technology has the potential to revolutionize voltage control in power systems. However, the state-of-the-art digital twin method suffers from low computational and sampling efficiency, which hinders its applications. To address this issue, we propose a \underline{G}umbel-\underline{C}onsistency \underline{D}igital  \underline{T}win (GC-DT) method that enhances voltage control with improved computational and sampling efficiency. 
First, the proposed method incorporates a Gumbel-based strategy improvement that leverages the Gumbel-top trick to enhance non-repetitive sampling actions and reduce the reliance on Monte Carlo Tree Search simulations, thereby improving computational efficiency. Second, a consistency loss function aligns predicted hidden states with actual hidden states in the latent space, which increases both prediction accuracy and sampling efficiency. 
 Experiments  on IEEE 123-bus, 34-bus, and 13-bus systems demonstrate that the proposed GC-DT outperforms the state-of-the-art DT method in both computational and sampling efficiency.
\end{abstract}


\section{Introduction}

Voltage control is a fundamental research problem, which directly determines the safe operation of power systems \cite{voltage control}. With the rapid growth in power demand, the load structure of the grid has become increasingly complex, and dynamic fluctuations in load have become more pronounced \cite{electrical load increase, yushuailiss}. Furthermore, the large-scale integration of distributed energy sources, such as photovoltaic and wind energy, has increased the volatility of energy production, making voltage more susceptible to fluctuations \cite{distributed energy,csj}. These developments urgently call for  flexible and efficient voltage control methods to adapt to the variability and complexity of modern distribution networks.

The methods to solve the voltage control problem can be divided into two categories: i) model-driven methods; and ii) data-driven methods. To be specific, model-driven  methods rely on constructing physical models to design control strategies, such as  droop control and model predictive control (MPC). Droop control adjusts reactive power output automatically to maintain voltage and frequency stability by responding to deviations in voltage and frequency. For instance, a droop control-based voltage regulation strategy was proposed in \cite{Droop control 1}, which improves voltage regulation of direct current (DC) distribution networks under uncertain conditions. An integral model predictive control strategy based on distributed online system recognition for voltage and frequency regulation  was proposed in \cite{distributed online system}, which can adapt to constantly changing system conditions.
With the increasing complexity of power systems and the rapid advancements in big data and artificial intelligence, voltage control methods in distribution grid systems have gradually transitioned from traditional model-driven strategies to data-driven approaches. Data-driven voltage control methods analyze historical data, real-time data, and environmental interaction information to learn system behavior patterns, eliminating the need for precise physical models. This approach offers greater flexibility in adapting to the complex dynamics of power systems. These methods are primarily divided into deep learning-based and reinforcement learning-based approaches.
Deep neural networks are employed to address complex voltage control problems by learning the nonlinear relationships within the power grid, thereby enhancing control accuracy \cite{Deep 0}. A real-time joint central and local voltage/reactive power control strategy based on deep learning and hierarchical optimization was proposed in \cite{Deep 1}. However, these approaches typically lack dynamic optimization capabilities and often require large amounts of labeled data for training. Reinforcement learning methods adjust control strategies by interacting with the environment and responding to reward signals to achieve optimal voltage control outcomes, without the need for labeled data \cite{RL}. Zhang et al. \cite{RL 1} embedded the voltage and reactive power optimization problem of the distribution system into a deep Q-network framework, enabling effective voltage regulation and power loss reduction. Additionally, a decentralized voltage control approach based on a partially observable Markov decision process was proposed in \cite{RL 5}, which optimized voltage stability and reduced network losses using multi-agent reinforcement learning algorithms and graph convolutional networks.
While these data-driven methods eliminate the need for physical models in voltage control, they often lack a well-defined multi-step exploration space for planning, which may  reduce the control performance, e.g., low mean reward.

To address this issue, digital twin (DT) technology offers a promising solution. A DT creates a real-time virtual model that mirrors a physical system through a digital representation. By simulating, analyzing, and optimizing this virtual model, the performance of the physical system can be better understood and predicted, allowing control strategies to be validated and optimized without directly impacting the actual system. 
For instance, the concept of DT was employed to solve the problem of power generation prediction for photovoltaics in \cite{DT 0}, which achieves high accuracy. An extra adjustment tool, the DT was employed to interact with the voltage control system \cite{DT 1}, in order to ensure stable operation of power systems. Indeed, voltage control is a complex control problem, which requires greater emphasis on accurately capturing the intrinsic dynamic behavior within its digital representation. In our recent study, we proposed a novel DT modeling approach that utilizes a self-learning model to digitally replicate dynamic characteristics, capturing the high-fidelity intrinsic behavior of electric vehicles \cite{my magazine}. While the method in \cite{my magazine} is general, we have two challenges when it is extended to the voltage control problem.   
Firstly, this DT method relies on the use of the Monte Carlo Tree Search (MCTS) to visit all possible actions. This process typically needs plenty of simulations to obtain the expected performance, resulting in low computational efficiency. Secondly, this DT method is built upon vast training data to obtain a  high-fidelity model with low sampling efficiency.  The solutions to solve the voltage problem in distribution grid systems  require high computational and sampling efficiency to support dynamic environments, which calls for new solutions. 


To address these two issues, the concept of Gumbel-top  distribution \cite{gumbel} and consistency-loss model \cite{consistency} may provide a promising solution. With this inspiration, this paper proposes a \underline{G}umbel-\underline{C}onsistency \underline{DT} modeling method, GC-DT for short, to solve the voltage control problem. 
The main cotributions are summarized as follows: 

1) A Gumbel-based policy improvement is employed, which leverages the use of  Gumbel-top distribution to enhance non-repetitive sampling actions. This approach enables DT models to acquire sufficient data for training robust strategies with fewer MCTS simulations, resulting in improved computational efficiency. 

2) A consistency loss function is incorporated to guide the training of the DT model, enforcing the predicted hidden states to align with the real hidden states in the latent space. This approach improves sampling efficiency by ensuring greater accuracy in the model's predictions. 

3) Experiments on IEEE 123-bus, 34-bus, and 13-bus systems shows that the proposed GC-DT outperforms the state-of-the-art DT method to solve the voltage control problem in terms of control performance, computational
efficiency, and sampling efficiency.


\section{Problem Formulation}

\subsection{Markov Decision Process}

The Markov Decision Process (MDP) is a mathematical framework used for modeling stochastic decision-making processes. It  indicates that the future state of the system depends solely on the current state and the action taken, with no dependence on prior states. 
An MDP is represented as a four-tuple $M = (S, A, P, R)$, where \(S\) is the state space, \(A\) is the action space, \(P\) is the transition function, and \(R\) is the reward function. The agent observes the current state \( s_t \in S \). Guided by the policy, the agent then takes action \( a_{t+1} \in A \), subsequently receiving an updated observation \( s_{t+1} \) and a reward \( u_{t+1} = R(s_t, a_{t+1}) \). In sequential decision-making tasks, such as voltage control, the goal is to achieve maximum return \( \sum_{t=0}^{T} \gamma_t u_t \), where \( \gamma_t \) is the discount factor at time \( t \).


\subsection{Voltage Control}
The voltage control system is modeled as a finite-horizon MDP with 
\( H = 24 \) time steps, representing the daily frequency of voltage control operations, with actions executed once per hour\cite{powergym}. The observation and action spaces are products of discrete and
continuous variables. Discrete variables correspond to the physical switching constraints of independent devices, such as the operation or disconnection of equipment and adjustments to voltage transformer tap settings. Continuous variables arise from devices like batteries, where actions such as the charge and discharge power are normalized to a bounded continuous range. The observed states associated with the voltage control problem are defined as:
\begin{align}\label{eq1}
s_t =[ s_t^{vol}, s_t^{cap}, s_t^{reg}, s_t^{soc}, s_t^{dis}], 
\end{align}
where $s_t^{vol}=\{s_t^{vol}(i,p)|i\in I,p\in 1,2,3\}$ is the set of bus voltages at time $t$,  $s_t^{vol}(i,p)$ represent the voltage at bus $i$ in phase $p$, and $I$ is the total number of buses; \(s_t^{cap}=\{s_t^{cap}(i)|i \in I_{cap}\}\) is 
the set of states of capacitors at time $t$, \(s_t^{cap}(i)\) represents the capacitor at bus $i$ (connection or disconnection), and \(I_{cap}\) is the number of buses with capacitors; \(s_t^{reg}=\{s_t^{reg}(i,p)|i \in I_{reg},p\in 1,2,3\}\) is the set of tap of regulators at time $t$, \(s_t^{reg}(i,p)\) represent the taps at bus $i$ in phase $p$, and \(I_{reg}\) is the number of buses with  regulators; \(s_t^{soc}=\{s_t^{soc}(i)|i \in I_{batt}\}\) and \(s_t^{dis}=\{s_t^{dis}(i)|i \in I_{batt}\}\) are the sets of batteries at time $t$, \(s_t^{soc}(i)\) and \(s_t^{dis}(i)\) represent the battery's state-of-charge (SOC) and its charge/discharge power at bus $i$, and \(I_{batt}\) is the number of  buses with batteries. 


The actions include:
\begin{align}\label{eq2}
    a_t = [a_t^{cap},a_t^{reg},a_t^{batt}],
\end{align}
where \(a_t^{cap}=\{a_t^{cap}(i)|i \in I_{cap}\}\) is the set of actions taken by capacitors at time \(t\), indicating their connection or disconnection; \(a_t^{reg}=\{a_t^{reg}(i)|i \in I_{reg}\}\) is the set of actions taken by regulators at time \(t\), indicating the adjustment of transformer tap positions; \(a_t^{batt}=\{a_t^{batt}(i)|i \in I_{batt}\}\)  is the set of actions taken by batteries at time \(t\), indicating their charging or discharging power. When multiple devices of the same type exist in the system, each operates independently. The action function aggregates the actions of all devices. The action space for batteries is classified into continuous and discrete types. Details of the observation and action spaces are provided in Table \ref{tab1}.
\renewcommand{\arraystretch}{1.1}
\begin{table}[htbp]
\caption{Observation and Action space}
\begin{center}
\begin{tabular}{cc|cc}
\hline
\textbf{Observation} & \textbf{Range} & \textbf{Action} & \textbf{Range} \\
\hline
Bus voltage & [0.8,1.2] & Capacitor& {0,1} \\
Capacitor & {0,1} & Regulator tap& {0,...,32} \\
Regulator tap & {0,...,32} & Discharge disc.& {0,...,32} \\
State of charge & [0,1] & Discharge cont.& [-1,1] \\
Discharge power & [-1,1] & &  \\
\hline
\end{tabular}
\label{tab1}
\end{center}
\end{table}

The reward function $u_t$, including a power loss item $f_{power} (s_{t+1})$, a voltage penalty item $f_{volt}(s_{t+1})$, and a control error item $f_{ctrl}  ( s_t,s_{t+1},t)$, are given by:
\begin{align}\label{eq3}
\nonumber
u_t =&-f_{power} (s_{t+1}) -f_{volt}(s_{t+1})\\
& -f_{ctrl}  ( s_t,s_{t+1},t),  
\end{align}
where \( t \in [0, \ldots, H - 1] \).

The power loss item is defined as:
\begin{align}\label{eq4}
f_{power}(s_{t+1}) =w_{power} \frac{PowerLoss ( s_{t+1} )}{TotalPower( s_{t+1} )},
\end{align}
where \(w_{power}\) is the weight for power losses; \(PowerLoss(s_{t+1})\) and \(TotalPower(s_{t+1})\) are the power loss and total power at time \(t+1\), respectively. 


The voltage penalty item is defined as:
\begin{align}\label{eq5}
\nonumber
f_{volt}(s_{t+1})=&\sum_{n\in N}\left(\max _{p} s_{t+1}^{vol}(n,p)-\overline{s}_{t+1}^{vol}(n,p)\right)_{+}\\
&+\left(\underline{s}_{t+1}^{vol}(n,p)-\min _{p} s_{t+1}^{vol}\left(n,p\right)\right)_{+}, 
\end{align}
\noindent where (\(\cdot\))+ is a shorthand for max(\(\cdot\),0); The superscript ``$\underline{\rule{8pt}{0ex}}$'' and the subscript ``$\overline{\rule{8pt}{0ex}}$'' represent the lower and upper bounds, respectively.

The  control error item is defined as:
\begin{align}\label{eq6}
\nonumber
f_{ctrl} &\left ( s,s_{t+1} ,t \right ) =\sum_{i\in I_{cap}}^{} w_{cap} \left | s_{t}^{cap}(i) -s_{t+1}^{cap}(i) \right | \\
\nonumber
&+\sum_{i\in I_{reg},p=1,2,3}^{} w_{reg} \left | s_{t}^{reg}(i,p) -s_{t+1}^{reg}(i,p) \right |  \\
&+\sum_{i\in I_{batt}}^{}\bigg{(}w_{dis} 
\frac{(s_{t+1}^{dis}(i))_{+} }{\overline{s}_t^{dis}(i)  } \\
\nonumber
&+w_{soc}\mathbb{I}_{t=H}\left | s_{t}^{soc}(i) -s_{0}^{soc}(i)  \right | \bigg{)},
\end{align}
where $w_{cap}$, $w_{reg}$, $w_{dis}$, and $w_{soc}$ are the penalty coefficients for control errors of capacitors, regulators, discharge/charge of batteries, and SOC  of batteries, respectively; $\mathbb{I}_{t=H}$ indicates the final time step. The control error item is set to effectively reduce the frequency of policy changes, minimize wear on control devices, and prevent ineffective responses caused by rapid equipment switching.


\section{Methodology}
The core concept of a DT model \cite{my magazine} is to create an abstract MDP and use it to make planning ahead. Specifically, the DT is composed of a transformation model, a dynamic model, and a prediction model, each of which is implemented using neural networks. The MCTS is employed to find the best action by combining with the three models. However, on the one hand, the implementation of MCTS requires to visit all actions. This leads to huge computing resources, resulting in reduced  computing  efficiency. On the other hand, to obtain a high-fidelity DT model, the original DT modeling method \cite{my magazine} requires extensive training with low sampling efficiency.  In real-world voltage control problems, high computational and sampling efficiency are crucial for energy conservation and stable operation.  
To address these efficiency challenges, the proposed GC-DT employs a Gumbel-based policy improvement strategy to enhance computational efficiency and uses a consistency loss to improve sampling efficiency.

\subsection{Gumbel-based Policy Improvement}\label{AA}
We introduce a Gumbel-based policy to achieve improvements in both action sampling and the strategy of action selection. Specifically, for the root node, we 
first employ a Gumbel canvas to generate $k$ Gumbel variables, apply strategy the \(\psi\) to perform non-replacement sampling, and ultimately  obtain \(m\) actions. Among these actions, the sequential halving method is used to quickly identify the action with the highest \(G\) (defined in Eq. \eqref{eq10}), which is the Gumbel-top sampling trick. 

The  Gumbel-max distribution is defined as:
\begin{align}\label{eq7}
\left ( g\in \mathbb{R}^{k}  \right ) &\sim Gumbel\left ( 0 \right ), \\
A&=\underset{a}{\arg \max} \left ( g\left ( a \right ) +logits\left ( a \right )  \right ), \label{eq8}
\end{align}
where \(logits(a)\) represents the logit of action \(a\).  To sample $A$
 from the Gumbel distribution, we first generate a vector of 
$k$ Gumbel variables and then take the operation of
${\arg \max} $. 

This Gumbel-max trick can be extended to sample 
$m$ actions without replacement by selecting the top 
$m$ actions. This is referred to as the Gumbel-top trick, defined as follows:
\begin{align}\label{eq9}
A_{\mu}=\underset{a\not\in{A_1,...A_{\mu-1}}}{\arg \max} \left ( g\left ( a \right ) +logits\left ( a \right )  \right ),
\end{align}
where $\mu \in 1, 2, \cdots, m$. We select $m$ 
top-actions, denoted as $Top(A_{\mu}, m)$.



In each simulation \(n \in \{1,...,N\}\), where \(N\) is the total number of simulations, the agent selects actions \(A_{n} \in \{0,...,k-1\}\) and visits the action to observe the Q-value as \(q(A_{n})\). Then, the agent find $m$ top actions with the highest $(g(a)+\log its(a))$ as $\text{Top}_{A_m}$, where $m \leq k$ is the number of actions sampled without replacement. Finally, the best actions $A_{n+1}$ are selected by using sequential halving with $n$ simulations to compare the highest $G$ from the $\text{Top}_{A_m}$, and the $G$ is defined as:
\begin{align}\label{eq10}
G=(g(a)+\log its(a)+\sigma(q(a)) ),
\end{align}
where \(\sigma\) is a monotonically increasing transformation; \(q(a)\) is the Q-value of the action \(a\). Moreover, the sequential halving method is used to efficiently identify the action with the highest $G$, which accelerates the selection of high-$G$ actions in order to improve computational efficiency.

The objective is to maximize the Q-value from a special latest action set \(A_{n+1}\). It means that we aim to maximize the expected value of \(q(A_{n+1})\), denoted as \(\mathbb{E}[q(A_{n+1})]\).
The following formula demonstrates that the $G$ trick provides a sufficiently good sample:
\begin{align}
\nonumber
&q\left(\underset{a \in \text{Top}_{A_m}}{\arg \max }(g(a)+\operatorname{logits}(a)+\sigma(q(a)))\right) \\
&\geq q\left(\underset{a \in \text{Top}_{A_m}}{\arg \max }(g(a)+\operatorname{logits}(a))\right). \label{eq11}
\end{align}
 
This inequality holds for any Gumbel \( g \), and therefore also holds for the expected value:
\(\mathbb{E}\left[q\left(A_{n+1}\right)\right] \geq \mathbb{E}_{a\sim \pi } \left[q\left(A_{n}\right)\right]\). Here, \(\mathbb{E}_{a\sim \pi }[q(A_{n})]\) represents the expected Q-value based on the action selection policy \(\psi\) in the \(n\)th simulation.

In the case of multiple simulations, some actions may be visited more than once. Therefore, using the empirical mean \(\hat{q}(a)\) instead of \(q(a)\) provides a more accurate estimate.
For a concrete formulation of \(\sigma\), the following approach is adopted:
\begin{align}\label{eq12}
\sigma(\hat{q}(a))=\left(c_{\text {visit }}+\max _b N(b)\right) c_{\text {scale }} \hat{q}(a),
\end{align}
where \(\max _b N(b)\) is the visit count of the most visited action. \(c_{\text{visit}}\) and \(c_{\text{scale}}\) are constants representing the number of visits and the scale value, respectively. 

After the search, the \(A_{n+1} \) predicted by the policy \(\psi( A_{n}) \) and the \(\hat{q}(a)\) for the visited actions are provided to improve the policy network. A simple policy loss is defined as \( L_{simple} \left ( \psi \right )=-\log \psi(A_{n+1}) \). 
Using the complete Q-values, an improved policy is constructed as:
\begin{align}\label{eq13}
\operatorname{completed\; Q}(a)= \begin{cases}\hat{q}(a) & \text { if } N(a)>0 \\ v_\psi & \text { otherwise }\end{cases},
\end{align}
where \( v_\psi=\sum_{a}^{} \psi \left ( a \right ) q\left ( a \right )\) is the estimated value for actions that have not been visited; \(\psi(a)\) presents  the probability of the policy network prediction for action \(a\).
  
Although the exact \( v_\psi \) is not available in practice, an approximate value from the value network \(\hat{v} _{\psi }\) is used instead. With the complete Q-values in place, an improved policy is constructed as follows:
\begin{align}\label{eq14}
\psi^{\prime}=\operatorname{softmax}(\operatorname{logits}+\sigma(\operatorname{completedQ})),
\end{align}
where \( \sigma \)  represents a monotonically increasing transformation.

\begin{algorithm}
\LinesNumbered
  \KwIn{Number of simulations n, round maximum time H, number of actions k, predictor logits from a policy network \(\psi\): \(logits \in \mathbb{R}^{k}\)}
  \KwOut{ (\(s_{t},\pi_{t},a_{t},u_{t},s_{t+1}\))}
  Initialization\;
  \While{t=1,2,3...in H}{
    \While{$n$\(\leq\)N}{
    Child node selection based on the current root node\;
    \eIf{Current node is a root node}{
    Observer $s_t$ from environment\;
   Generate $k$  Gumbel variables $\leftarrow$ Eq. \eqref{eq7}\;
    $\text{Top}_{A_m}$$\leftarrow$ Eq. \eqref{eq9}\;
    Compute  \(G\)\ $\leftarrow$ Eq. \eqref{eq10};\\
    \(A_{n+1}\) = \(argmax_{a \in \text{Top}_{A_m}}(G)\)\;
    }{
    Deterministic actions $\leftarrow$ Eq. \eqref{eq15}\;
    \Return \(A_{n+1}\)
    }{ 
    $\psi' \leftarrow$ Eq. \eqref{eq14}\;
    \(\psi \leftarrow\) Eq. \eqref{eq16}\;
    $n=(n+1)$\;
    }
    { }
    }{\Return $\pi_{t}$\;
    $a_t\leftarrow \pi_{t}$\;
    Obtain $s_{t+1}$ and $u_t$ from environment\;
    \(t=t+1\)\;
    \Return(\(s_{t},\pi_{t},a_{t},u_{t},s_{t+1}\))
    }}
\caption{ Gumbel-based policy improvement in CG-DT}
\end{algorithm}

For action selection at non-root nodes in the search tree, an improved policy \(\psi'\) is constructed by using the complete Q-values. Since sampling at non-root nodes introduces unnecessary variance to the estimated Q-values, a deterministic action selection method was designed, as follows:
\begin{align}\label{eq15}
\underset{a}{\arg \max }\left(\psi^{\prime}(a)-\frac{N(a)}{1+\sum_b N(b)}\right),
\end{align}
where \(N(b)\) and \(N(a)\) are  the visit counts of actions $a$ and $b$, respectively. 

After constructing the  improved network, it can be refined into a new policy network \( \psi \). The loss function trains on all actions, not just the action \(A_{n+1}\), as defined by:
\begin{align}\label{eq16}
L_{\text {completed }}(\psi)=\operatorname{KL}\left(\psi^{\prime}, \psi\right),
\end{align}
where \(\text{KL}\) represents the Kullback–Leibler divergence between the two probability distributions.

By using the improved policy for action selection at the root node and employing deterministic selection at non-root nodes, this method maintains a high expected value for action selection strategies, enhancing both sampling performance and efficiency in complex environments.  The pseudocode of 
Gumbel-based policy improvement in CG-DT 
 is provided in Algorithm 1.

\subsection{Model Improvements}
The improvement of the neural network model involves the addition of a consistency loss model. In \cite{my magazine}, training relies solely on rewards, values, and policies. Due to their scalar nature, these metrics sometimes fail to provide sufficient training information. This issue is exacerbated when rewards are sparse or guiding values are inaccurate. 
Furthermore, an accurate and good strategy largely depends on the accuracy of environmental samples and sample utilization, so it is crucial to establish precise indicators to construct accurate strategies.

To achieve accurate policies, the output \( \hat{s}_{t+1} \) from the dynamic model is expected to align with the actual hidden state \( \hat{s}'_{t+1} \), which  corresponds to the output derived from the transformation model using the next environment observation \( s_{t+1} \). 
By using the actual hidden state \( \hat{s}'_{t+1} \)  to improve the model's prediction of the next hidden state 
\(\hat{s}_{t+1} \), more training signals are generated, enhancing the training process. This consistency loss function helps reduce the number of required training steps and increases the model's overall efficiency. The consistency loss function is defined as:
\begin{align}\label{17}
L_{\text{consistency}}(\hat{s}'_{t+1} ,\hat{s}_{t+1})=\mathcal{L}(P_1(\hat{s}'_{t+1}), P_2(\hat{s}_{t+1})),
\end{align}

\noindent where $P_1(\hat{s}'_{t+1})$ is the output of the neural network $P_1(\cdot)$ by taking $\hat{s}'_{t+1}$ as input. $P_2(\hat{s}_{t+1})$ is the output of the neural network $P_2(\cdot)$ by taking $\hat{s}_{t+1}$ as input. \(\mathcal{L}\) represents the Cross-Entropy loss function.

The overall  loss function is given by:
\begin{align}\label{eq18}
\nonumber
    L_t(\theta)=&\lambda_1L(u_t, r_t)+\lambda_2 L(\pi_t, p_t)+\lambda_3 L(z_t, v_t) \\
+&\lambda_4 L_{\text {consistency}}\left(\hat{s}'_{t+1}, \hat{s}_{t+1}\right)+c\|\theta\|^2,
\end{align}
where $\lambda_1$, $\lambda_2$, $\lambda_3$, and $\lambda_4$ are the weights in the loss function; \(c\) is a positive coefficient. The first three items in Eq. \eqref{eq18} are the same as the counterparts in  \cite{my magazine}.

Three neural network models are jointly trained to obtain the model of GC-DT. To be specific,
at time step \( t \), the task begins with the initial environment observation \(s_{t}\), which serves as the input to the transformation model, yielding the hidden state \(\hat{s}_{t}\). The dynamic model then updates the hidden state iteratively by taking the previous hidden state \( \hat{s}_{t}\) and the current action \( a_{t}\) to derive the hypothesized reward \( r_{t}\) and the new hidden state \( \hat{s}_{t+1}\). Each \( \hat{s}_{t+1} \) is then fed into the prediction model to compute the predicted policy  \( p_{t} \) and the value function \( v_{t} \). At the end of each episode, the transformation model, dynamic model, and prediction model are trained via backpropagation by minimizing Eq. \eqref{eq18}.

\subsection{ GC-DT based Voltage Control}

After constructing the model of GC-DT, it can be used to solve the voltage control for power
systems. 
Initially, the power system gathers current and historical states $s_t$, creating a stacked observation at each time step $t$, which is then fed into the GC-DT model.  $s_t$ passed into the transformation network as input and subsequently converted into the initial digital states $\hat{s}_t^0$. Taking $\hat{s}_t^0$ as the root node, the Gumbel-based policy improvement performs $N$ simulations for future $k$-step searching as detailed in Algorithm 1. This process allows the power system to generate a variety of explorations and to select the optimal actions in the digital space. Ultimately, each component receives and executes the selected actions in the physical power systems.

\section{Experiments}

Experiments are conducted in three environments, including IEEE 123-bus, IEEE 34-bus,  and IEEE 13-bus systems, following \cite{powergym}. The proposed GC-DT and the original and state-of-the-art DT  method \cite{my magazine} are compared for solving the voltage control problem with the same settings.   

\subsection{Analysis of Efficiency}

\begin{figure*}[htbp]
\centering
\includegraphics[scale=.55]{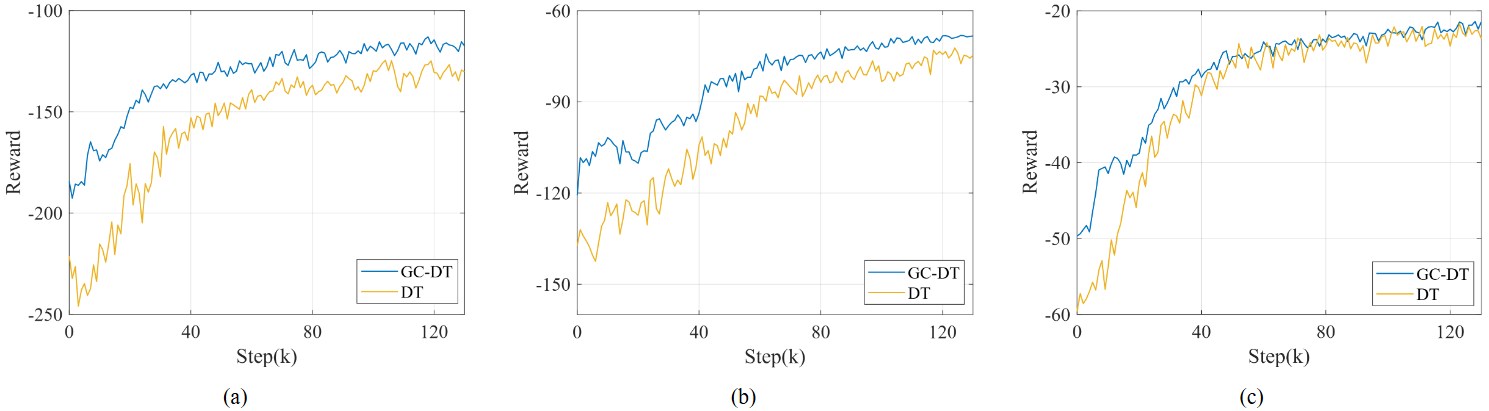}
\caption{Comparisons of the GC-DT and the original DT  with $n=16$ in IEEE system. (a) IEEE 123-bus. (b) IEEE 34-bus. c) IEEE 13-bus.}
\label{fig1}
\end{figure*}

The average awards by performing the proposed GC-DT and the original DT method \cite{my magazine} across all the three systems with  $n=16$ are shown in Fig. \ref{fig1}, respectively. 
 As observed, in the IEEE 123-bus and IEEE 34-bus systems, the GC-DT achieves significantly higher average rewards compared to the original DT. In the IEEE 13-bus system, the average rewards for both methods are relatively similar, likely due to its less complex nature compared to the IEEE 123-bus and 34-bus systems. Therefore, both GC-DT and original DT yield higher average rewards. However, the GC-DT requires considerably fewer steps to achieve a stable average reward in the IEEE 123-bus, 34-bus, and 13-bus systems compared to the original DT. This suggests that the GC-DT offers improvements in both sampling and computational efficiency, especially in complex environments, where its superior performance is more evident.

\subsection{Comparison  Studies}


\begin{table*}[t]
\centering
\caption{Comparison of the GC-DT and the original DT  with different simulations $n$ in IEEE system}
\scalebox{1.1}{
\begin{tabular}{c|cc|cc|cc|cc|cc|cc}
\hline
Model & \multicolumn{4}{c|}{IEEE 123-bus}  & \multicolumn{4}{c|}{IEEE 34-bus} & \multicolumn{4}{c}{IEEE 13-bus}\\
\hline
Simulations & \multicolumn{2}{c}{n=4} & \multicolumn{2}{c|}{n=16} & \multicolumn{2}{c}{n=4} & \multicolumn{2}{c|}{n=16} & \multicolumn{2}{c}{n=4} & \multicolumn{2}{c}{n=16}\\
\hline
Steps (k) & GC-DT & DT & GC-DT & DT & GC-DT & DT & GC-DT & DT & GC-DT & DT & GC-DT & DT\\ 
\hline
20& -160.1 & -174.1 & -147.9 & -175.6 & -119.3& -126.5 & -110.2 & -127.3 & -43.6& -47.2& -38.8& -42.6 \\

40& -152.7 & -176.1 & -131.8 & -152.9 & -109.2& -125.5 & -93.8 & -104.2 & -35.8& -40.1& -28.8& -31.2\\

60& -151.2 & -175.7 & -126.2 & -139.1 & -95.1& -122.8 & -76.7 & -88.1 & -29.4& -31.7& -24.1& -25.6\\

80& -149.2 & -176.7 & -126.1 & -136.9 & -83.5& -118.7 & -73.6 & -83.6 & -27.5& -28.9& -24.1& -24.6\\
\hline
\end{tabular}
}
\label{table2}
\end{table*}

Table \ref{table2} shows the average rewards during training by performing the two methods across the three  systems. It can be observed that as the decrease of $n$ (from 16 to 4), the original DT fails to yield a satisfactory result, whereas GC-DT still continues to achieve good outcomes. This indicates that the GC-DT is capable of offering high quality voltage control (high reward)  with fewer simulations, owing to the improved  computational efficiency compared to the original DT method.

\begin{table}[htbp]
\centering
\caption{The results of TS and TE}
\begin{tabular}{c|cc|cc|cc}
\hline
\multirow{2}{*}{\textbf{Model}} &\multicolumn{2}{c|}{\textbf{IEEE 123-bus}}& \multicolumn{2}{c|}{\textbf{IEEE 34-bus}}&\multicolumn{2}{c}{\textbf{IEEE 13-bus}}\\
\cline{2-7}
 & TS (s) & TE (s) & TS (s) & TE (s) & TS (s) & TE (s) \\
\hline
GC-DT & 0.54 & 68.16 & 0.47 & 83.21 & 0.48 & 64.37 \\
DT & 0.69 & 86.12 & 0.64 & 97.78 & 0.66 & 74.68 \\
\hline
\end{tabular}
\label{tab3}
\end{table}

Moreover, the average time of each step (TS) and  the average time of sampling of each episode (TE) are summarized in Table \ref{tab3}. It can be observed that the GC-DT algorithm increases: i) TS by 22\% and TE by 21\%, respectively in the IEEE 123-bus system; ii) TS by 27\% and TE by 15\%, respectively in the IEEE 34-bus system; and iii) TS by 27\% and TE by 14\%, respectively in the IEEE 34-bus system. This further provides evidence that the proposed GC-DT outperforms the original DT in the aspects of both computational efficiency and sampling efficiency.





\section{Conclusion and Future Work}

This paper proposes a new GC-DT method to solve the voltage control problem for power systems. The proposed method  integrates two key innovations: a Gumbel-based sample policy improvement and a consistency loss function. With those components, the proposed GC-DT is capable of significantly improving both computational and sampling efficiency, resulting in reduced a need for extensive MCTS simulations while maintaining high accuracy of the leaned model. Experiments on three IEEE 123-bus, 34-bus, and 13-bus systems verify the effectiveness of the proposed method.

\end{document}